\documentclass[letterpaper]{article} 
\usepackage{aaai24}  
\usepackage{times}  
\usepackage{helvet}  
\usepackage{courier}  
\usepackage[hyphens]{url}  
\usepackage{graphicx} 
\usepackage{natbib}  
\usepackage{caption} 
\DeclareCaptionStyle{ruled}{labelfont=normalfont,labelsep=colon,strut=off} 
\usepackage{latexsym}
\usepackage{url}
\usepackage{graphicx}
\usepackage{multirow}
\usepackage{makecell}
\usepackage{lscape}
\usepackage{bm}
\usepackage[ruled,boxed,commentsnumbered,linesnumbered]{algorithm2e}
\usepackage{algpseudocode}
\usepackage{framed}
\usepackage{color}
\usepackage{xcolor}
\usepackage{booktabs,colortbl}
\usepackage{subfigure}
\usepackage{multirow} 

\usepackage{amssymb}
\usepackage{amsmath}
\usepackage{soul}
\usepackage{tabularx}
\usepackage{booktabs}
\usepackage{utfsym}

\newcommand{\eg}{\emph{e.g., }}

\newcommand{\baby}{ABSA-ESA\xspace}

\title{Aspect-Based Sentiment Analysis with Explicit Sentiment Augmentations}
\author{
    Jihong Ouyang\textsuperscript{\rm 1,\rm 2}\equalcontrib,
    Zhiyao Yang\textsuperscript{\rm 1,\rm 2}\equalcontrib,
    Silong Liang\textsuperscript{\rm 1,\rm 2},
    Bing Wang\textsuperscript{\rm 1,\rm 2},
    Yimeng Wang\textsuperscript{\rm 1},
    Ximing Li\textsuperscript{\rm 1,\rm 2}\thanks{Corresponding author}
}
\affiliations{
    \textsuperscript{\rm 1}College of Computer Science and Technology, Jilin University, China\\
    \textsuperscript{\rm 2}Key Laboratory of Symbolic Computation and Knowledge Engineering of MOE, Jilin University, China\\
    \{ouyj@, zhiyaoy20@mails., liangsl23@mails.\}jlu.edu.cn, \{wangbing1416, wangyimeng116, liximing86\}@gmail.com
}

\begin{document}

\maketitle

\begin{abstract}
    Aspect-based sentiment analysis (ABSA), a fine-grained sentiment classification task, has received much attention recently. Many works investigate sentiment information through opinion words, such as ``good'' and ``bad''. However, \textbf{implicit sentiment} data widely exists in the ABSA dataset, whose sentiment polarity is hard to determine due to the lack of distinct opinion words. To deal with implicit sentiment, this paper proposes an ABSA method that integrates explicit sentiment augmentations (\baby) to add more sentiment clues. We propose an ABSA-specific explicit sentiment generation method to create such augmentations. Specifically, we post-train T5 by rule-based data and employ three strategies to constrain the sentiment polarity and aspect term of the generated augmentations. We employ Syntax Distance Weighting and Unlikelihood Contrastive Regularization in the training procedure to guide the model to generate the explicit opinion words with the same polarity as the input sentence. Meanwhile, we utilize the Constrained Beam Search to ensure the augmentations are aspect-related. We test \baby on two ABSA benchmarks. The results show that \baby outperforms the SOTA baselines on implicit and explicit sentiment accuracy.
\end{abstract}

\section{Introduction}

\begin{figure}
    \centering
    \includegraphics[width=0.47\textwidth]{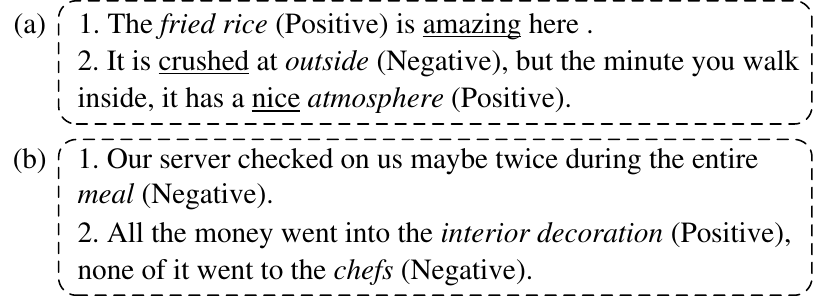}
    \captionsetup{font={small}} 
    \caption{Examples of the explicit sentiment sentences (a) vs. implicit sentiment sentences (b) from Rest.14 dataset. The words with \underline{underline} are the opinion words.}
    \label{Example}
\end{figure}

\textbf{A}spect-\textbf{b}ased \textbf{S}entiment \textbf{A}nalysis (\textbf{ABSA}) aims to induce predictive models over manually annotated sentences to identify the sentiment polarity towards each specific aspect term \cite{wang2022contrastive,li2022sk2}. Taking the second sentence in Fig.~\ref{Example} (a) as an example, the task aims to automatically identify the sentiment polarities of its aspect terms ``\textit{{outside}}'' (\texttt{Negative}) and ``\textit{{atmosphere}}'' (\texttt{Positive}) potentially with the corresponding opinion words ``\underline{crushed}'' and ``\underline{nice}''. Due to its popularity, ABSA has been widely applied in many real-world scenarios, and accordingly, it is one of the most significant tasks in the natural language processing community \cite{yang2023s3,ouyang2023unsupervised}.

To handle the task of ABSA, many studies have been investigated during the past decade. Broadly speaking, the focus of recent work is on how to generate more discriminative representations for aspect terms to enhance the identification performance of sentiment polarity. Some early studies generate strong aspect term representations by directly employing deep neural encoders, such as LSTM \cite{tang2016effective, wang2016attention, cheng2017aspect} and pre-trained language models \cite{xu2020understanding,dai2021does}. Beyond them, to further link the aspect terms and opinion words, some studies build dependency trees of sentences and then generate aspect term representations by employing graph convolution networks (GCN) \cite{sun2019aspect,wang2020relational,chen2020inducing,li2021dual}. 

The success of the GCN-based approach underscores the pivotal role that opinion words play in the realm of ABSA. However, recent research has highlighted a complex scenario characterized by a lack of distinct opinion words, termed "implicit sentiment" \cite{li2021learning, wang2022causal}. To delve into this phenomenon, we select four examples from the Rest.14 to compare the implicit and explicit sentiment sentences. In the context of Fig.\ref{Example}(a), the sentiment is discernible due to distinct opinion words. In contrast, as shown in Fig.\ref{Example}(b), unraveling the sentiment associated with aspect terms such as "meal," "interior decoration," and "chefs" is challenging. Implicit sentiment is a prevalent occurrence within ABSA datasets and it is hard to deal with \cite{li2021learning}. 

To tackle the challenge mentioned above, in the paper, we design a novel \textbf{ABSA} method by integrating \textbf{E}xplicit \textbf{S}entiment \textbf{A}ugmentations (\textbf{\baby}). Such augmentations provide more sentiment clues for predicting sentiment polarity. We add them after the corresponding input sentence, forming new ABSA training data. To obtain the augmentations, we design an \textbf{ABSA-specific explicit sentiment generation} method. We aim to generate the sentences \textbf{explicitly} conveying the \textbf{same sentiment polarity} as their corresponding input sentences, targeting \textbf{the same (or similar) aspect terms}. We post-train the generation model T5 \cite{raffel2020exploring} by the rule-based data selected in the ABSA dataset, making the generated augmentations comply with the above requirements. Furthermore, we introduce three strategies to confine the generated augmentations about their sentiment polarity and aspect terms. Specifically, in the training procedure, we employ the \textbf{Syntax Distance Weighting} and \textbf{Unlikelihood Contrastive Regularization} to lead the model to generate explicit opinion words with the same polarity as the input sentence. Subsequently, when engendering the augmentations, we employ the \textbf{Constrained Beam Search} to ensure the augmentations are aspect-related.

To sum up, our contributions can be listed as follows:
\begin{itemize}
    \item We propose a novel ABSA framework named \baby, which focuses on solving the implicit sentiment issue by generating explicit sentiment augmentations.
    \item We propose an ABSA-specific explicit sentiment generation method that generates augmentations with distinct opinion words for specific aspect terms.
    \item Empirical results on two ABSA benchmarks show that \baby outperforms other methods on both explicit and implicit accuracy.
\end{itemize}

\section{Related Work}
\subsection{Aspect-based Sentiment Analysis}
Aspect-Based Sentiment Analysis (ABSA) methods primarily focus on integrating sentiment information from contextual words into aspect terms. In earlier approaches, this was often achieved by utilizing LSTM or Bi-LSTM as encoders \cite{tang2016effective, wang2016attention, cheng2017aspect}. Consequently, recent advancements have embraced the Attention mechanism as the preferred encoder \cite{tang2016effective, wang2016attention, cheng2017aspect}. Notably, leveraging pre-trained language models has emerged as the prevailing trend in ABSA \cite{xu2020understanding, dai2021does}. Furthermore, to establish stronger connections between aspect terms and opinion words, numerous studies have delved into constructing dependency trees within sentences and refining aspect term representations using Graph Convolutional Networks (GCNs) \cite{sun2019aspect, wang2020relational, li2021dual}.

Concurrently, alongside developing robust encoders, researchers have explored the enrichment of training data to provide external sentiment information for the model \cite{he2019interactive, wang2022contrastive, yang2023s3}. These additional data often lack fine-grained annotations and necessitate subsequent data processing. Addressing this, this paper integrates ABSA-specific augmentations into ABSA models, bypassing the need for extensive reprocessing.

\subsection{Implicit Sentiment Analysis}
Implicit sentiment classification, a pivotal subfield within sentiment analysis, was pioneered by \citet{liu2012sentiment}, drawing significant scholarly interest. Initial works revolved around implicit sentiment at the sentence level \cite{deng2014joint, choi2017coarse, zhou2021implicit, xu2022kc}. Recent endeavors have shifted towards tackling implicit aspect-based sentiment classification \cite{li2021learning, wang2022causal, fei2023reasoning}. A prevailing approach involves incorporating external knowledge to capture sentiment expression patterns. For instance, \citet{xu2022kc} integrates external sentiment-related knowledge into sentence features, enhancing the model's sentiment comprehension. Similarly, \citet{li2021learning} employs a post-training strategy with BERT, leveraging contrastive learning on expansive sentiment-annotated corpora. \baby utilizes the data generated by the model instead of obtaining external knowledge.

\subsection{Data Augmentation}
Within NLP, the data augmentation technique has gained substantial traction to expand the pool of available training instances. This approach finds widespread application across diverse domains, including text classification \citep{wu2022text, liu2022dynamic,ouyang2022weakly}, neural machine translation \citep{lam2022sample, kambhatla2022cipherdaug, gao2019soft}, and text generation \citep{bi2021data, xu2021augnlg}. Notably, recent strides in ABSA have similarly leveraged data augmentation \citep{chen2022unsupervised, wang2022contrastive, hsu2021semantics}. However, their augmentation techniques tend to be relatively simple, \eg token replacement, masked aspect prediction, and polarity reversal, limiting the semantic diversity of the enhanced samples. The augmentation method in this paper is based on the language model, which generates augmentations with rich sentiment information. 

\begin{figure*}[t]
    \centering
    \includegraphics[scale=1.35]{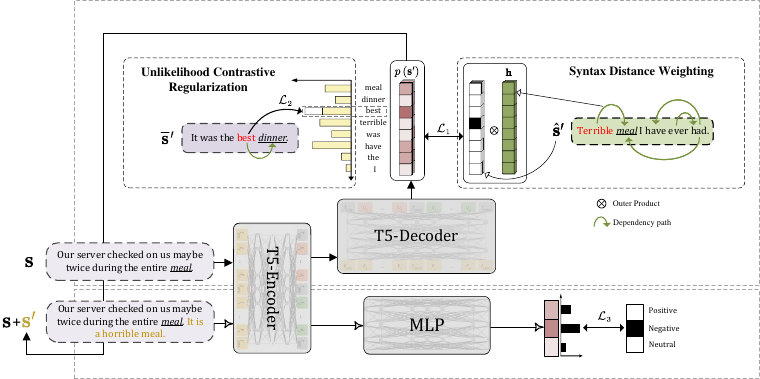}
    \caption{Overall framework of \baby.}
    \label{framework}
\end{figure*}

\section{Our Proposed \baby Model}
In this section, we introduce the proposed ABSA method named \textbf{\baby}.

\subsection{Overall Framework}
Generally speaking, ABSA methods take the review sentence $\mathbf{s} = \{s_{j}\}_{j=1}^M$ and its corresponding aspect term $\mathbf{a}=\{a_j\}_{j=1}^{|\mathbf{a}|}$ as the model input, $M$ denotes the length of all sentence. And output the sentiment polarity $y \in \mathcal{Y} = \{\texttt{Positive}, \texttt{Negative}, \texttt{Neutral}\}$ for $\mathbf{a}$. To deal with the sentences containing implicit sentiment, we extend this paradigm by introducing an augmented sentence $\mathbf{s}'$ following the initial input $\mathbf{s}$. This augmented sentence contains explicit sentiment tied to the aspect term $\mathbf{a}$. For clarity, we present the comprehensive framework of \baby in Figure~\ref{framework}.

To generate the augmented sentence $\mathbf{s}'$, we propose an ABSA-specific explicit sentiment generation method. We post-train T5 by utilizing $\hat{\mathbf{s}}'$ as generation targets selected from the dataset. $\hat{\mathbf{s}}'$ has the same (or similar) aspect terms and sentiment as $\mathbf{s}$ while also incorporating explicit sentiment expressions. Additionally, we utilize three strategies to guide the generation concerning sentiment polarity and aspect terms. During the training phase, a Syntax Distance Weighting strategy is implemented to prioritize context words closest to the aspect term in the dependency parse. Furthermore, we also gather $\Bar{\mathbf{s}}'$, which has the opposite sentiment of $\mathbf{s}$, for Unlikelihood Contrastive Regularization. It instructs the model about undesirable word choices. 
When generating $\mathbf{s}'$, we employ Constrained Beam Search to ensure that the aspect term or its similar aspect is included in the augmentations and its context words are the most relevant to $\mathbf{a}$.

Next, we introduce the details of the ABSA-specific explicit sentiment generation method. 

\subsection{Training Data Collection}
To train the explicit sentiment generation model, the initial step is to gather the training data. Given an input sentence $\mathbf{s}$ and its corresponding aspect term $\mathbf{a}$, the generating target $\hat{\mathbf{s}}'$ must satisfy the following rules:
\begin{itemize}
    \item The target sentence should incorporate \textbf{the same (or similar) aspect term} as the input sentence.
    \item The target sentence should exhibit \textbf{identical sentiment polarity} to the input data.
    \item The target sentence must \textbf{contain explicit sentiment expressions}. 
\end{itemize}

To obtain the target sentence $\hat{\mathbf{s}}'$ that satisfies the given rules, we begin by aggregating all aspect terms in dataset $\mathcal{D}$ to construct the aspect term set $\mathcal{A}=\{\bar{\mathbf{a}}_i\}_{i=1}^{|\mathcal{A}|}$. Each aspect term $\bar{\mathbf{a}}_i$ is associated with a representation $r_{\bar{\mathbf{a}}_i}$\footnote{In the case where an aspect term $\mathbf{a}$ consists of multiple words ($\mathbf{a} = {w_1, w_2,\cdots, w_{|\mathbf{a}_i|}}$), we calculate the average word embedding $\bar{r}$ as the representation for that aspect.} acquired by consulting the GloVe embedding table \cite{pennington2014glove}. Utilizing these representations $\mathcal{R} =\{ r_{\bar{\mathbf{a}}_i}\}_{i=1}^{|\mathcal{A}|}$, we formulate a similarity matrix $\mathbf{C}=\{c_{ij}\}_{|\mathcal{A}|\times |\mathcal{A}|}$, where $c_{ij}$ represents the similarity between aspect terms $\bar{\mathbf{a}}_i$ and $\bar{\mathbf{a}}_j$. $c_{ij}$ is computed by the cosine distance:
\begin{equation}
c_{ij} = \cos(r_{\bar{\mathbf{a}}_i}, r_{\bar{\mathbf{a}}_j}).
\end{equation}
With the similarity matrix $\mathbf{C}$ available, we proceed to the selection of $\hat{\mathbf{s}}'$. According to \citet{li2021learning}, the dataset $\mathcal{D}$ can be divided into explicit subset $\mathcal{D}_e$ and implicit subset $\mathcal{D}_i$. As sentences in $\mathcal{D}_e$ contain explicit sentiment expressions, we choose the $\hat{\mathbf{s}}'$ from this subset to fulfill the third rule above.

We first select $k_c$ aspect terms from $\mathcal{R}$, which are most similar to $\mathbf{a}$, thereby forming the set $\mathcal{A}'$. Subsequently, we extract sentences from $\mathcal{D}_e$ containing aspect terms from $\mathcal{A}'$ and share the same sentiment as $\mathbf{s}$. This forms the candidate target sentence set $\hat{\mathcal{S}_t}$.  From this set, we randomly choose a target sentence $\mathbf{\hat{s}}'\in \hat{\mathcal{S}_t}$ to generate training data $(\mathbf{s},\mathbf{\hat{s}}',\mathbf{a})$ with the input sentence and the corresponding aspect term. This process is iterated for all input sentences $\mathbf{s} \in \mathcal{D}$, resulting the final training dataset $\hat{\mathcal{D}}=\{(\mathbf{s}_i,\mathbf{\hat{s}}'_i,\mathbf{a}_i)\}_{i=1}^N$. Afterward, we begin to post-train T5 by $\hat{\mathcal{D}}$.

\subsection{Sentiment Polarity Constraints}
While our training pairs have been thoughtfully chosen, it's important to acknowledge that ABSA datasets often contain sentences with multiple aspect words and contrasting sentiment polarities. Take the following training data as an example:

\textbf{Input sentence}: I have been told to choose the \textit{{food}} (\texttt{Positive}) in this restaurant several times.

\textbf{Target sentence}: \underline{Bad} \textit{service} (\texttt{Negative}) while the \textit{{food}} (\texttt{Positive}) here in our dinner is \underline{worthy} of being recommended.

In this instance, the target sentence involves two aspect terms, ``service'' and ``food,'' associating with opposing sentiment polarities. Directly training the model with such pairs may lead the model to generate "bad" with "food" for positive polarity. To navigate this challenge, we integrate two constraints in training \baby: Syntax Distance Weighting (SDW) and Unlikelihood Contrastive Regularization (UCR).

\paragraph{Syntax Distance Weighting.} Syntax relationships are widely utilized in ABSA to establish a closer connection between aspect terms and their corresponding opinion words \cite{wang2020relational,li2021dual,hou2021graph,zhou2021closer}. To channel \baby's focus towards generating explicit opinion words, we introduce varying learning weights to words within target sentences based on their syntax distance from the aspect term. The shorter the syntax distance between a context word and the aspect term, the higher the assigned learning weight. To start, we employ a dependency parser to construct the dependency tree of the target sentence $\hat{\mathbf{s}}'_i$. Subsequently, we calculate syntax distances between the aspect term and the context words, culminating in the formation of a distance vector $\mathbf{d}_i=\{d_{ij}\}_{j=1}^M$. Finally, syntax distance weights are computed using the ensuing formula:

\begin{equation}
\mathbf{h}_i = \mathbf{1}-\mathrm{Softmax}(\mathbf{d}_i),
\end{equation}
where $\mathbf{1}$ denotes the matrix of ones with the shape same as $\mathbf{d}_i$. Then, we train the generation model with $(\mathbf{s}_i,\mathbf{\hat{s}}'_i, \mathbf{a}_i)$ by minimize the following loss:
\begin{equation}
\label{eq1}
\mathcal{L}_1 = -\sum_{j=1}^M h_{ij} \log p(\hat{s}'_{ij}|\mathbf{a}_i;\mathbf{s}_i; \mathbf{\hat{s}}'_{i;\leq j}),
\end{equation}
where $\mathbf{\hat{s}}'_{i;\leq j}$ denotes the sub-sentence from $\hat{s}'_{i1}$ to $\hat{s}'_{ij}$, and $;$ denotes the concatenate operation.

\paragraph{Unlikelihood Contrastive Regularization.}
To mitigate the adverse effect of unrelated words in the target sentence, we select the negative target sentence $\Bar{\mathbf{s}}_i'$ from $\mathcal{D}_e$ that shares the same aspect term as the input sentence $\mathbf{s}_i$ yet has the opposite sentiment\footnote{Both \texttt{Positive} and \texttt{Negative} can be the opposite sentiment to \texttt{Neutral}.}. We obtain the dependency tree of $\Bar{\mathbf{s}}_i$ and calculate the distance vector same as the SDW approach. Subsequently, we generate a negative word set $\Bar{\mathbf{s}}'^-=\{\Bar{s}'^-_i\}_{i=1}^{k_n}$ by selecting the top-$k_n$ nearest words to the aspect term. Importantly, note that $\Bar{\mathbf{s}}_i' \cap \hat{\mathbf{s}}_i' = \varnothing$, ensuring words from $\hat{\mathbf{s}}_i'$ are excluded when choosing negative words. Next, we minimize the Unlikelihood Contrastive Regularization loss as follows:
\begin{equation}
    \mathcal{L}_2 = -\sum_{j=1}^M \log\frac{p(\hat{s}'_{ij})}{p(\hat{s}'_{ij})+\sum_{l=1}^k p(\Bar{s}'^-_k)},
\end{equation}

\begin{algorithm}[t]
    \small
    
    \SetKwInput{KwInput}{Input}     
    \SetKwInput{KwOutput}{Output} 
     \linespread{1}\selectfont
        \SetAlgoLined
        \caption{t-th Step Constrained Beam Search}
        \label{CBS}
        \KwInput{Candidate sub-sentence $\mathcal{G}_{t-1}$, Model $\mathcal{F}$ Aspect term $\mathbf{a}$ and its similar aspects $\mathcal{A}'$}
        \KwOutput{Generated Sentence or  $\mathcal{G}_{t}$}
        \BlankLine
        \For{$\mathbf{g} \in \mathcal{G}_{t-1}$}{
        $p\leftarrow$ Compute next word probability of $\mathcal{F}(\mathbf{g})$;\\
        $\mathcal{W} \leftarrow$ Select V words with top probability according $p$;\\
        $\mathcal{W} \leftarrow \mathcal{W} \cup \mathbf{a} \cup \mathcal{A}'$ ;\\
            \For{$w \in \mathcal{W}$}{ 
               $\mathbf{g}\leftarrow$ Add word $w$ to $\mathbf{g}$.\\
                $\hat{\mathcal{G}}^{t}\leftarrow$ add $\mathbf{g}$ to $\hat{\mathcal{G}}^{t}$   
            }
        }
    
        \eIf{$\hat{\mathbf{g}}\in \hat{\mathcal{G}}^{t}$ contains \textless/s\textgreater and the aspect term}{
            $\mathbf{s}' \leftarrow$ Output $\hat{\mathbf{g}}$ as the generated sentences.
        }{
            $\mathcal{G}_t \leftarrow$ Randomly select $V$ sub-sentences in $\hat{\mathcal{G}}^{t}$.\\
            Do the next constrained beam search step.
        }
    \end{algorithm}

\subsection{Aspect Term Constraint}
Text generation with the language model is a Hidden-Markov-like process where each generated word is related to its front words. Inspired by this, we aim to have the generated sentence incorporate the input aspect term, resulting in context words that are intrinsically related to the aspect. Specifically, we utilize the Constrained Beam Search (CBS), which generates a set of sub-sentence $\mathcal{G}=\{\mathbf{g}_i\}_{i=1}^V$ at each generating step instead of just outputting one. At the $t$-th step, the previous sub-sentences $\mathcal{G}^{t-1}$ serve as input, and the subsequent word is generated for every sub-sentence, thereby forming candidate sub-sentences  $\hat{\mathcal{G}}^{t}$:
\begin{equation}
    \hat{\mathcal{G}}^{t} = \{(\mathbf{g};w)|\mathbf{g}\in \mathcal{G}^{t-1},w\in \mathcal{W}\},\\
\end{equation}
\begin{equation}
    \label{gw}
    \mathcal{W} = \text{argTopZ}(p(\mathbf{g}^{t-1})) \;\cup \; \Bar{\mathcal{A}}'
\end{equation}
where $\Bar{\mathcal{A}}'=\mathbf{a} \;\cup\; \mathcal{A}'$ is the aspect terms set that \baby need to generate. $;$ is the concatenate operation and argTopZ($\cdot$) output $z$ words with the highest generation probability. The CBS process concludes when any sentence $\hat{\mathbf{g}} \in \hat{\mathcal{G}}^{t}$ contains the aspect term and the ending symbol \textless/s\textgreater. CBS output $\hat{\mathbf{g}}$ as the generated sentence $\mathbf{s}'$. By contrast, if no sentence fulfills the above requirement, we randomly select $V$ sentence to form $\mathcal{G}^{t}$ for the subsequent CBS step. For clarity, Algorithm~\ref{CBS} provides the $i$-th CBS step process.

\subsection{Overall Objective}
Upon obtaining the augmented sentence $\mathbf{s}'_i$, we integrate it with the input sentence $\mathbf{s}_i$, yielding novel training triplets  $(\mathbf{s}^c_i,\mathbf{a}_i,y_i)$, where $\mathbf{s}^c_i=\{s_{i1},s_{i2},\cdots,s'_{i1},\cdots,s'_{i|\mathbf{s'}|}\}$ represents the combined sentence. We then input $\mathbf{s}_i^c$ and the corresponding aspect term $\mathbf{a}_i$ into the T5-Encoder to obtain the aspect representation $\mathbf{H}_i^a$. Then we input $\mathbf{H}_i^a$ into a multi-layer perception (MLP) to obtain the sentiment prediction $\hat{y}_i$:
\begin{equation}
    \hat{y}_i = \textrm{MLP}(T5_e(\mathbf{s}_i^c, \mathbf{a}_i))),
\end{equation}
Finally, we employ a Cross-Entropy loss to guide \baby in sentiment prediction:
\begin{equation}
    \mathcal{L}_3 = \ell_{\mathrm{CE}}(y_i,\hat{y}_i),
\end{equation}
Overall, we train \baby by minimizing the following objective:
\begin{equation}
    \mathcal{L} = \mathcal{L}_1 + \mathcal{L}_2 + \mathcal{L}_3.
\end{equation}

\begin{table}[t]
    \caption{Statistics of the Rest.14 and Lap.14 dataset. The ``\textbf{Implicit}'' column denotes the number of sentences with implicit sentiment expression.}
    \centering
    \footnotesize
    \label{statistics}
    \renewcommand{\heavyrulewidth}{0.13em}
    \renewcommand{\lightrulewidth}{0.07em}
    \renewcommand{\cmidrulewidth}{0.05em}
    \renewcommand\arraystretch{1.2}
    \newcolumntype{P}[1]{>{\centering\arraybackslash}p{#1}}
    \begin{tabular}{p{2cm}*{4}{P{1cm}}}
        \toprule
        \textbf{Dataset}                  & \textbf{Positive} & \textbf{Negative} & \textbf{Neutral}  & \textbf{Implicit} \\
        \midrule
        \textbf{Lap.14}$_{\textrm{train}}$         & 987               & 460               & 866               & 715               \\
        \textbf{Lap.14}$_{\textrm{test}}$          & 341               & 169               & 128               & 174               \\
        \textbf{Rest.14}$_{\textrm{train}}$        & 2164              & 633               & 805               & 1030              \\
        \textbf{Rest.14}$_{\textrm{test}}$         & 728               & 196               & 196               & 267               \\
        \bottomrule
    \end{tabular}
    \end{table}

\begin{table*}[h]
\caption{Main results of \baby and each baselines. The result with \textbf{Bold} is the best result. The results with $\flat$ are from \cite{wang2022causal} and the results with $\natural$ are from \cite{wang2022contrastive}. The results produced by ourselves are labeled with $\dagger$. All, ESE and ISE denote the result obtained by using all data, explicit data and implicit data, respectively. The subscripts $A$ and $F$ represent the accuracy and macro-F1 score, respectively.}
\label{mainresult1}
\centering
\footnotesize
\renewcommand{\heavyrulewidth}{0.13em}
\renewcommand{\lightrulewidth}{0.07em}
\renewcommand{\cmidrulewidth}{0.05em}
\renewcommand\arraystretch{1.2}
\newcolumntype{P}[1]{>{\centering\arraybackslash}p{#1}}
\begin{tabular}{p{0.3cm}p{5.5cm}*{8}{P{0.9cm}}}
\toprule
    &\multirow{2}{*}{\textbf{Model}} & \multicolumn{4}{c}{\textbf{Rest.14}} & \multicolumn{4}{c}{\textbf{Lap.14}}              \\
    \cmidrule(lr){3-6} \cmidrule(lr){7-10} 
    &                                & All$_A$             & All$_F$                & ESE$_A$           & ISE$_A$           & All$_A$          & All$_F$            & ESE$_A$           & ISE$_A$      \\
    \midrule
    \multirow{7}{*}{\rotatebox{90}{\textbf{BERT}$_{bace}$}}
    &BERT-$base$(110M)$^\flat$               & 83.57            & 77.16             & 89.21         & 65.54         & 78.22         & 73.45         & 81.47         & 69.71     \\
    &BT$^\natural$ \cite{fan2021beyond}      & 86.47            & 80.29             & 93.37         & 64.41         & 79.67         & 75.79         & 82.60         & 72.21      \\
    &BERTAsp \cite{li2021learning}           & 85.80            & 78.95             & 92.73         & 63.67         & 78.53         & 74.07         & 82.34         & 68.39      \\
    &EDA$^\natural$ \cite{wei2019eda}        & 86.42            & 79.63             & 92.83         & 65.91         & 79.59         & 75.79         & 83.15         & 70.44      \\
    &PT$^\flat$ \cite{xu2019bert}            & 84.95            & 76.96             & 91.26         & 64.79         & 78.07         & 75.08         & 81.47         & 71.27      \\
    &ADA$^\flat$ \cite{rietzler2020adapt}    & 87.14            & 80.05             & \textbf{94.14}& 65.92      & 78.96         & 74.18         & 82.76         & 70.11      \\
    &R-GAT$^\flat$ \cite{wang2020relational} & 86.60            & 81.35             & 92.73         & 67.79         & 78.21         & 74.07         & 82.44         & 72.99      \\ 
    &CLEAN$^\flat$ \cite{wang2022causal}     & 87.05            & 81.40             & 92.50         & 69.66         & 80.41         & 77.25         & 81.21         & 78.29      \\
    \midrule    
    \multirow{4}{*}{\rotatebox{90}{\textbf{BERT}$_{large}$}}    
    &BERT-$large$(340M)$^\dagger$            & 86.89            & 79.62             & 93.45         & 66.06         & 79.77         & 76.64         & 82.73         & 72.02      \\
    &R-GAT$^\dagger$\cite{wang2020relational}& 86.79            & 80.17             & 93.77         & 68.55         & 79.62         & 75.95         & 81.22         & 75.42      \\
    &BERTAsp$^\dagger$ \cite{li2021learning} & 87.01            & 81.07             & 93.25         & 67.05         & 80.78         & 77.54         & 83.48         & 73.71      \\
    &ASA-WD \cite{tian2021enhancing}         & 86.88            & 80.92             & 92.62         & 68.53         & 80.41         & 77.38         & 83.41         & 72.57      \\
    \midrule    
    \multirow{7}{*}{\rotatebox{90}{\textbf{T5}$_{base}$}}   
    &T5-$base$(220M)$^\dagger$               & 86.60            & 79.24             & 93.08         & 65.92         & 80.25         & 76.13         & 83.17         & 72.60      \\
    &C$^3$DA$^\natural$ \cite{wang2022contrastive}& 86.93       & 81.23             & 93.59         & 65.54         & 80.61         & 77.11         & 82.68         & 73.57      \\
    &R-GAT$^\dagger$ \cite{wang2020relational}& 86.87           & 79.99             & 92.84         & 68.05         & 80.25         & 76.26         & 82.05         & 75.43      \\
    &T5Asp$^\dagger$ \cite{li2021learning}& 87.11               & 80.95             & 93.27         & 67.42         & 79.49         & 77.46         & 81.92         & 73.14      \\
    \cmidrule(lr){2-10}
    &\baby                           &\textbf{88.29} & \textbf{81.74} & 93.77 &\textbf{70.78} &\textbf{82.44} &\textbf{79.34} &\textbf{83.35}& \textbf{80.00}\\
 
    \bottomrule
\end{tabular}
\end{table*}

\section{Experiments}

\subsection{Dataset and Implementation Details}
In our experiments, we use the dataset released by \citet{li2021learning}, which has already been labeled with explicit and implicit sentiment. The statistics of the dataset are presented in Table~\ref{statistics}. In the training process, we set the learning rate as $5\times 10^{-5}$ (Restaurant) and $2\times 10^{-5}$ (Laptop). The batch size is 4 (Restaurant) and 8 (Laptop). We set the training epoch as 15 for both datasets. We set $k_c$=2 in data selection and $k_n$=4 for Unlikelihood Contrastive Regularization. For Constrained Beam Search, we select $V$=6 and $z$=3.

\subsection{Baselines}
\paragraph{Common ABSA Baselines.} PT \cite{xu2019bert} used BERT, which post-trained on several ABSA sub-tasks, as the encoder. ADA \cite{rietzler2020adapt} conducts ABSA by incorporating data from other domains. R-GAT \cite{wang2020relational} learns both aspect term embedding and dependency relation embedding to obtain more information. ASA-WD \cite{tian2021enhancing} uses the key-value memory network to explore the word dependencies.
\paragraph{Implicit ABSA Baselines.} BERTAsp \cite{li2021learning}\footnote{We select the results of BERTAsp with the version not trained with external data. And the results of its variants, such as T5Asp, are also obtained in the same way.} utilizes contrastive learning to post-train BERT for more implicit sentiment knowledge. CLEAN \cite{wang2022causal} explores implicit sentiment by studying the causal relations in each sentence. THOR \cite{fei2023reasoning} employs the chain-of-thought strategy to manually provide more sentiment clues for the implicit sentiment sentences.
\paragraph{Data Augmentation Baselines.} Back Translation (BT) translates the training sentence into another language and then translates it back. EDA \cite{wei2019eda} augments data at the token level by methods such as synonym replacement and random insertion. C$^3$DA \cite{wang2022contrastive} generates augmentations by replacing the aspect term and reversing the sentiment polarity. 

\begin{table}[t]
\caption{Results compared with THOR. All the baselines use Flan-T5 as the encoder. The results with $\heartsuit$ are obtained from \cite{fei2023reasoning}. The metrics are the same as Table~\ref{mainresult1}.}
\label{mainresult2}
\centering
\footnotesize
\renewcommand{\heavyrulewidth}{0.13em}
\renewcommand{\lightrulewidth}{0.07em}
\renewcommand{\cmidrulewidth}{0.05em}
\renewcommand\arraystretch{1.2}
\newcolumntype{P}[1]{>{\centering\arraybackslash}p{#1}}
\begin{tabular}{p{3.5cm}*{4}{P{0.7cm}}}
\toprule
    \multirow{2}{*}{\textbf{Model}} & \multicolumn{2}{c}{\textbf{Rest.14}} & \multicolumn{2}{c}{\textbf{Lap.14}}              \\
    \cmidrule(lr){2-3} \cmidrule(lr){4-5} 
                                        & All$_F$           & ISE$_F$           & All$_F$             & ISE$_F$       \\
    \midrule
    Flan-T5(250M)                       & 81.52             & 69.66         & 79.32          & 72.02     \\
    Flan-T5 Prompt $^\heartsuit$        & 81.50             & 70.91         & 79.02          & 76.40     \\
    THOR $^\heartsuit$\cite{fei2023reasoning}& 82.98             & 71.70         & 79.75          & 67.63     \\
    \cmidrule(lr){1-5}
    \baby$_{\textrm{Flan-T5}}$          & \textbf{83.79}    & \textbf{73.76}& \textbf{81.78} & \textbf{77.91} \\
    \bottomrule
\end{tabular}
\end{table}

\begin{table}[t]
\caption{Ablation study of \baby. w/o denotes the version without the specific component. The metrics are the same as Table~\ref{mainresult1}.}
\label{ablation}
\centering
\footnotesize
\renewcommand{\heavyrulewidth}{0.13em}
\renewcommand{\lightrulewidth}{0.07em}
\renewcommand{\cmidrulewidth}{0.05em}
\renewcommand\arraystretch{1.2}
\newcolumntype{P}[1]{>{\centering\arraybackslash}p{#1}}
\begin{tabular}{p{3.5cm}*{4}{P{0.7cm}}}
\toprule
    \multirow{2}{*}{\textbf{Model}} & \multicolumn{2}{c}{\textbf{Rest.14}} & \multicolumn{2}{c}{\textbf{Lap.14}}              \\ \cmidrule(lr){2-3} \cmidrule(lr){4-5} 
                    & ESE$_A$           & ISE$_A$           & ESE$_A$           & ISE$_A$       \\ \midrule
    w/o SDW         &  92.78            & 68.16             & 82.64             & 77.58         \\
    w/o UCR         &  93.53            & 69.28             & 83.00             & 79.31         \\
    w/o SDW and UCR &  92.67            & 68.39             & 82.41             & 77.90         \\
    w/o CBS         &  93.36            & 67.81             & 82.06             & 78.65         \\\midrule
    Full \baby      & \textbf{93.77}    & \textbf{70.78}    & \textbf{83.35}    & \textbf{80.00}\\    
    \bottomrule
\end{tabular}
\end{table}

\subsection{Main Results}
The main results are in Table~\ref{mainresult1} and Table~\ref{mainresult2}, organized based on the backbone employed by each baseline. In general, \baby consistently demonstrates superior performance compared to the baselines across all evaluation metrics. Remarkably, \baby significantly enhances the classification accuracy of implicit sentiment data in contrast to the state-of-the-art CLEAN baseline (1.12$\uparrow$ on Rest.14 and 1.71$\uparrow$ on Lap.14). Additionally, \baby exhibits notable advancements in the classification accuracy of explicit sentiment data, (average 1.27$\uparrow$ on Rest.14 and 2.14$\uparrow$ on Lap.14). In the subsequent analysis, we delve into a detailed comparison of results across various baseline categories.

\paragraph{Compared with Implicit ABSA Baselines.} 
Our initial focus is comparing \baby against implicit ABSA baselines closest to \baby. We first focus on the BERTAsp and CLEAN. Even though these methods employ cutting-edge methods such as contrastive learning and causal reasoning, the results in Table~\ref{mainresult1} show that \baby has the best implicit accuracy. This indicates that generating explicit sentiment augmentations is a sample and effective way to deal with implicit ABSA.

Subsequently, we compare \baby with THOR. Although both \baby and THOR share the core notion of enriching the input sentence with additional sentiment cues, there exists a fundamental distinction between them. THOR employs predefined prompts to guide model responses, whereas \baby directly generates explicit sentiment tied to the given aspect. The results in Table~\ref{mainresult2} illustrate that \baby outperforms THOR on both datasets, particularly excelling in the F1-score of implicit data. It's noteworthy that on the Lap.14 dataset, THOR's implicit data F1-score trails behind that of the basic prompt model (67.63 vs. 76.40). This underscores the vulnerability of methods built on the chain-of-thought strategy, where outcomes are heavily reliant on manually designed prompts. In contrast, \baby's strength lies in its autonomous generation of explicit sentiment augmentations aligned with the provided aspect terms.

\begin{table*}[ht]
\caption{Case study of \baby. The sentence in \textbf{Bold} is the augmented sentence, and the word in \textcolor{red}{red} color is the generated explicit sentiment expression.}
\label{case}
\footnotesize
\centering
\renewcommand{\heavyrulewidth}{0.13em}
\renewcommand{\lightrulewidth}{0.07em}
\renewcommand{\cmidrulewidth}{0.05em}
\renewcommand\arraystretch{1.2}
\newcolumntype{P}[1]{>{\centering\arraybackslash}p{#1}}
\begin{tabular}{P{1cm}p{11cm}P{1.8cm}P{2.2cm}}
\toprule
    \textbf{Aspect} & \makecell[c]{\textbf{Sentence}} & \textbf{Prediction} & \textbf{Ground Truth} \\  
    \midrule
     \multirow{2}{*}[-1.9ex]{hostess} & \multicolumn{1}{m{11cm}}{I have tried to make reservations, but both times, the \textit{hostess} didn't have my name.} & \texttt{Neutral} & \multirow{2}{*}[-1.5ex]{\texttt{Negative}} \\  
    \cmidrule(lr){2-3}
     &  \multicolumn{1}{m{11cm}}{I have tried to make reservations, but both times, the \textit{hostess} didn't have my name. \textbf{Hostess service was \textcolor{red}{awful}.}} &\texttt{Negative} &  \\ 
    \midrule
     \multirow{2}{*}[-1.9ex]{waitress} & \multicolumn{1}{m{11cm}}{The \textit{waitress} suggested glasses of wine that went very well with the food.} & \texttt{Neutral} & \multirow{2}{*}[-1.5ex]{\texttt{Positive}} \\  
    \cmidrule(lr){2-3}
     &  \multicolumn{1}{m{11cm}}{The \textit{waitress} suggested glasses of wine that went very well with the food. \textbf{The waitress was very \textcolor{red}{nice and attentive}.}} &\texttt{Positive} &  \\ 
     \midrule
     \multirow{2}{*}[-1.9ex]{features} &  \multicolumn{1}{m{11cm}}{I was looking for a Mac which is portable and has all the \textit{features} that I was looking for.} & \texttt{Neutral} & \multirow{2}{*}[-1.5ex]{\texttt{Positive}} \\  
    \cmidrule(lr){2-3}
     &  \multicolumn{1}{m{11cm}}{I was looking for a Mac which is portable and has all the \textit{features} that I was looking for. \textbf{I \textcolor{red}{love} the features and the design.}} & \texttt{Positive} &  \\ 
    \midrule
     \multirow{2}{*}[-1ex]{service} &  \multicolumn{1}{m{11cm}}{Going to bring it to service today.} & \texttt{Neutral} & \multirow{2}{*}[-1ex]{\texttt{Neutral}} \\  
    \cmidrule(lr){2-3}
     &  \multicolumn{1}{m{11cm}}{Going to bring it to service today. \textbf{I have not seen a problem with the service.}}& \texttt{Neutral} &  \\ 
    \bottomrule
\end{tabular}
\end{table*}

\paragraph{Compared with Common ABSA Baselines.} 
Reviewing Table~\ref{mainresult1}, it becomes apparent that the overall and explicit accuracy of the common ABSA methods and \baby are relatively similar. However, a more pronounced disparity emerges when analyzing the accuracy of implicit data. \baby outperforms these common ABSA methods by a considerable margin. This indicates that the most common ABSA methods struggle with implicit sentiment expression. Notably, among these methods, R-GAT presents a relatively competitive implicit performance by incorporating dependency relations within sentences to foster closer connections between aspect terms and opinion words. Nonetheless, there remains a substantial gap compared to \baby's performance. 

\paragraph{Compared with Data Augmentation Baselines.} 
Analyzing Table~\ref{mainresult1},  \baby demonstrates a slight edge over the state-of-the-art data augmentation ABSA method (C$^3$DA) concerning overall F1 scores. However, the advantage becomes more pronounced when considering implicit sentiment, with \baby surpassing C$^3$DA by a significant margin. This divergence stems from \baby's direct inclusion of sentiment clues within the input sentence and its ability to generate aspect-related opinion words autonomously. In contrast, C$^3$DA's augmentation quality heavily relies on seed spans, which produce sentiment polarity inversion with the input sentence.

\subsection{Ablation Study}
In this section, we analyze the influence of the three strategies on the final results, as outlined in Table~\ref{ablation}. Notably, all strategies exhibit minimal impact on explicit sentiment data. Conversely, for implicit sentiment data, the omission of these strategies results in substantial drops in performance, particularly evident with the SDW strategy removed (2.62$\downarrow$ on Rest.14 and 2.42$\downarrow$ on Lap.14). Furthermore, a comparison between the version without SDW and the version without both strategies reveals no significant divergence in their results. This implies that relying on UCR offers limited enhancements to implicit accuracy and may even reduce model performance. Practical implementation suggests combining SDW and UCR for optimal outcomes.

\begin{table}[t]
\caption{Average prediction entropy of each model. All models are based on T5.}
\label{entorpy}
\centering
\footnotesize
\renewcommand{\heavyrulewidth}{0.13em}
\renewcommand{\lightrulewidth}{0.07em}
\renewcommand{\cmidrulewidth}{0.05em}
\renewcommand\arraystretch{1.2}
\newcolumntype{P}[1]{>{\centering\arraybackslash}p{#1}}
\begin{tabular}{p{2.5cm}*{4}{P{0.9cm}}}
\toprule
     \multirow{2}{*}{\textbf{Model}} & \multicolumn{2}{c}{\textbf{Rest.14}} & \multicolumn{2}{c}{\textbf{Lap.14}} \\ \cmidrule(lr){2-3} \cmidrule(lr){4-5}
     & All&ISE&All&ISE\\
    \midrule
    T5-base &  2.8936 & 2.5178 & 2.9123 & 3.0806 \\
    R-GAT   &  5.3242 & 5.4460 & 5.9510 & 5.5374 \\
    T5Asp   &  4.2844 & 4.5287 & 4.8858 & 4.9604 \\
    \midrule
    \baby       & \textbf{2.4267}& \textbf{2.3965}    & \textbf{2.5218}& \textbf{2.6315} \\

    \bottomrule
\end{tabular}
\end{table}

\subsection{Case Study}
In this section, we present some cases in Table~\ref{case} to show the effectiveness of \baby. We select totally four examples from two datasets. Firstly, it is evident that \baby adeptly generates explicit sentiment expressions (highlighted in red within the sentences) in line with the input sentences, incorporating aspect-related opinion words. For instance, the opinion word ``attentive'' matches the aspect term ``waitress'' very well. Secondly, a discernible pattern emerges from the model's predictions: for implicit sentiment data, the model consistently outputs a \texttt{Neutral} sentiment polarity. After the augmentation, the model's predictions align accurately with the ground truth sentiment. Thirdly, in scenarios with \texttt{Neutral} ground truth sentiment, \baby refrains from generating explicit sentiment. The augmentation for the final example has no opinion words, preserving the \baby's original prediction.

\subsection{Further Analysis}
In this section, we introduce a novel perspective employing average entropy to dissect the ABSA results. The average entropy of ABSA model predictions is calculated as follows:
\begin{equation}
    H(x)=-\frac{1}{n}\sum_{i=1}^n\sum_{c\in \mathcal{Y}} p_c(x_i)\log p_c(x_i),
\end{equation}
where $p_j(x_i)$ denotes the probability of the sentence $i$ on class $c$. As showcased in Table~\ref{entorpy}, \baby boasts the lowest average entropy under the same backbone (T5). This outcome suggests that not only does \baby accurately classify implicit sentiment data, but its classification tendencies are markedly distinct. This phenomenon arises primarily due to the integration of explicit sentiment augmentations, which endows sentences with pronounced sentiment inclinations, consequently enhancing discernibility. Furthermore, it is notable that both the overall and implicit entropy of \baby is lower than those of the T5-base. This suggests that the incorporation of explicit sentiment augmentations not only enhances implicit classification tendencies but also brings about benefits to explicit sentiment data.

\section{Conclusion}
This paper proposes an ABSA method to deal with implicit sentiment expression by integrating explicit sentiment augmentations (\baby). We design a novel ABSA-specific explicit sentiment generation method based on the T5 model and involve three strategies to constrain the generated augmentations on sentiment polarity and the aspect term. Specifically, Syntax Distance Weighting and Unlikelihood Contrastive Regularization are used to lead the model to generate explicit opinion words with the correct polarity. And Constrained Beam Search is used to ensure the augmentations are aspect-related. We test \baby on two ABSA benchmarks, and empirical results demonstrate that \baby can outperform existing ABSA baselines on both explicit and implicit sentiment accuracy.

\section*{Acknowledgements}
This work is supported by the project from the National Key R\&D Program of China (No.2021ZD0112500), the National Natural Science Foundation of China (NSFC) (No.62276113), the Project Funded by China Postdoctoral Science Foundation (No.2022M721321), the Energy Administration of Jilin Province (No.3D516L921421), and the Fundamental Research Funds for the Central Universities, JLU.

\bibliography{aaai24}

\end{document}